\documentclass{article}
\usepackage{spconf,amsmath,graphicx}
\usepackage{booktabs}
\usepackage[normalem]{ulem}
\useunder{\uline}{\ul}{}

\usepackage{subcaption} 

\title{Efficient keyword spotting using dilated convolutions and gating}
 \name{\begin{tabular}{c}Alice Coucke, Mohammed Chlieh, Thibault Gisselbrecht, David Leroy,\\ Mathieu Poumeyrol, Thibaut Lavril \end{tabular}}
\address{Snips, Paris, France}
\begin{document}
%
\maketitle
\begin{abstract}
\begin{sloppypar}
We explore the application of end-to-end stateless temporal modeling to small-footprint keyword spotting as opposed to recurrent networks that model long-term temporal dependencies using internal states. We propose a model inspired by the recent success of dilated convolutions in sequence modeling applications, allowing to train deeper architectures in resource-constrained configurations. Gated activations and residual connections are also added, following a similar configuration to WaveNet. In addition, we apply a custom target labeling  that back-propagates loss from specific frames of interest, therefore yielding higher accuracy and only requiring to detect the end of the keyword. Our experimental results show that our model outperforms a max-pooling loss trained recurrent neural network using LSTM cells, with a significant decrease in false rejection rate. The underlying dataset
-- ``Hey Snips'' utterances recorded by over 2.2K different speakers -- 
 has been made publicly available to establish an open reference for wake-word detection.
\end{sloppypar}

\end{abstract}


\section{Introduction}

\label{sec:intro}
Keyword spotting (KWS) aims at detecting a pre-defined keyword or set of keywords in a continuous stream of audio. In particular, wake-word detection is an increasingly important application of~KWS, used to initiate an interaction with a voice interface. In practice, such systems run on low-resource devices and listen continuously for a specific wake word. An effective on-device KWS therefore requires real-time response and high accuracy for a good user experience, while limiting memory footprint and computational cost.

Traditional approaches in keyword spotting tasks involve Hidden Markov Models (HMMs) for modeling both keyword and background \cite{rose1990hidden, wilpon1990automatic, wilpon1991improvements}. In recent years, Deep Neural Networks (DNNs) have proven to yield efficient small-footprint solutions, as shown first by the fully-connected networks introduced in \cite{chen2014small}. More advanced architectures have been successfully applied to KWS problems, such as Convolutional Neural Networks (CNNs) exploiting local dependencies \cite{sainath2015convolutional, zhang2017hello}. They have demonstrated efficiency in terms of inference speed and computational cost but fail at capturing large patterns with reasonably small models. Recent works have suggested RNN based keyword spotting using LSTM cells that can leverage longer temporal context using gating mechanism and internal states \cite{fernandez2007application, sun2016max, baljekar2014online}. However, because RNNs may suffer from state saturation when facing continuous input streams \cite{chang2018temporal}, their internal state needs to be periodically reset. 
 
In this work we focus on end-to-end stateless temporal modeling which can take advantage of a large context while limiting computation and avoiding saturation issues. By \textit{end-to-end} model, we mean a straight-forward model with a binary target that does not require a precise phoneme alignment beforehand. We explore an architecture based on a stack of dilated convolution layers, effectively operating on a broader scale than with standard convolutions while limiting model size. We further improve our solution with gated activations and residual skip-connections, inspired by the WaveNet style architecture explored previously for text-to-speech applications \cite{van2016wavenet} and voice activity detection \cite{chang2018temporal}, but never applied to KWS to our knowledge. In \cite{tang2017deep}, the authors explore Deep Residual Networks (ResNets) for KWS. ResNets differ from WaveNet models in that they do not leverage skip-connections and gating, and apply convolution kernels in the frequency domain, drastically increasing the computational cost.

In addition, the long-term dependency our model can capture is exploited by implementing a custom ``end-of-keyword'' target labeling, increasing the accuracy of our model. A max-pooling loss trained LSTM initialized with a cross-entropy pre-trained network is chosen as a baseline, as it is one of the most effective models taking advantage of longer temporal contexts \cite{sun2016max}. The rest of the paper is organized in two main parts. Section \ref{sec:model} describes the different components of our model as well as our labeling. Section \ref{sec:exp} focuses on the experimental setup and performance results obtained on a publicly available ``Hey Snips'' dataset\footnote{https://research.snips.ai/datasets/keyword-spotting}.

\section{Model Implementation}
\label{sec:model}

\subsection{System description}
\label{sec:descr}
\begin{sloppypar}
The acoustic features are 20-dimensional log-Mel filterbank energies (LFBEs), extracted from the input audio every 10ms over a window of 25ms. A binary target is used, see Section \ref{sec:eoh_labeling} for more details about labeling. During decoding, the system computes smoothed posteriors by averaging the output of a sliding context window containing $w_{smooth}$ frames, a parameter chosen after experimental tuning. \textit{End-to-end} models such as the one presented here do not require any post-processing step besides smoothing, as opposed to multi-class models such as \cite{chen2014small, sainath2015convolutional}. Indeed,  the system triggers when the smoothed keyword posterior exceeds a pre-defined threshold.
\end{sloppypar}

\subsection{Neural network architecture}
\begin{figure}[tb!]
    \centering
    \includegraphics[width=8.5cm]{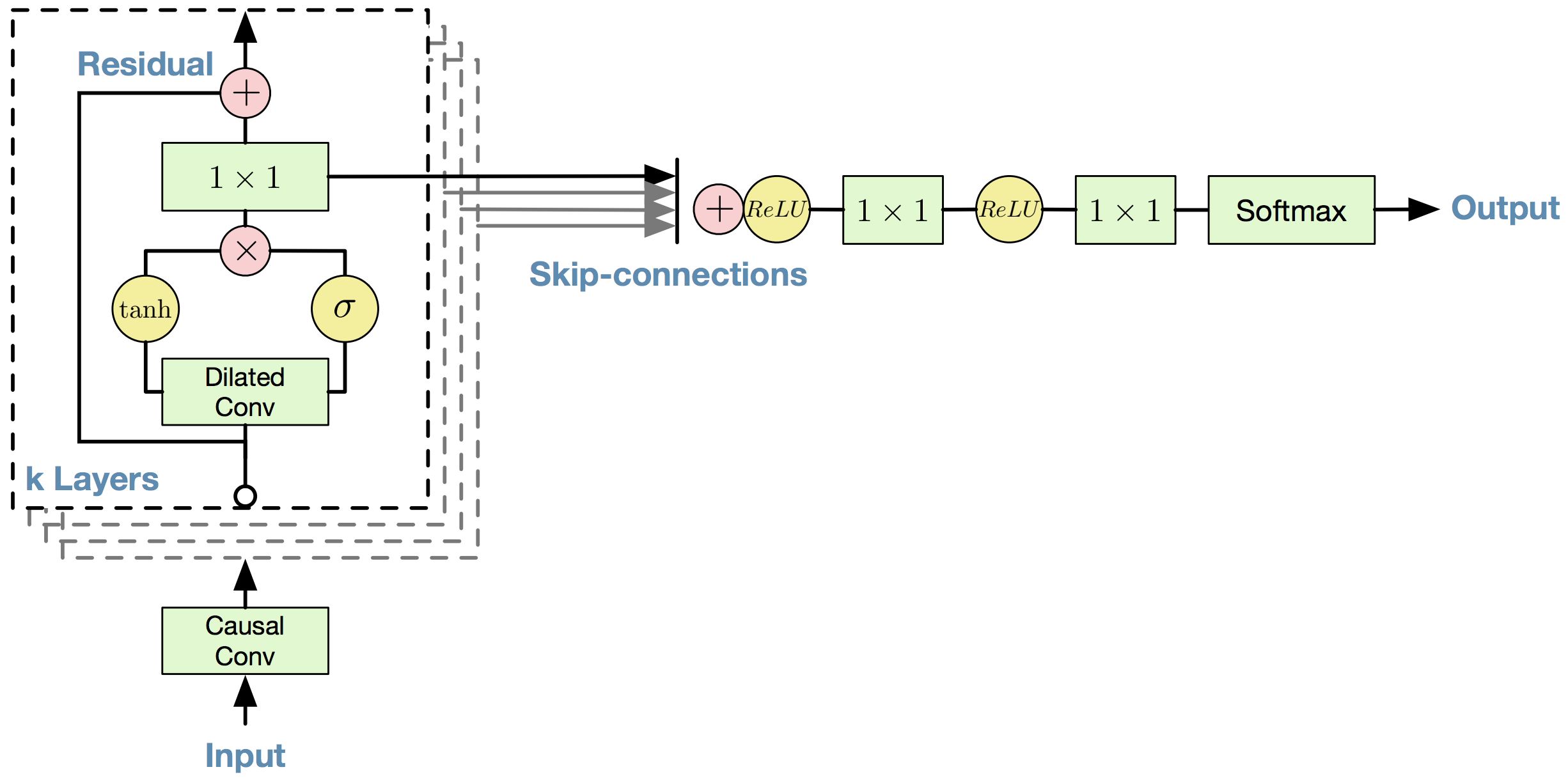}
    \caption{WaveNet architecture \cite{van2016wavenet}.}
    \label{fig:wavenet}
\end{figure}

WaveNet was initially proposed in \cite{van2016wavenet}, as a generative model for speech synthesis and other audio generation tasks. It consists in stacked causal convolution layers wrapped in a residual block with gated activation units as depicted in Figure~\ref{fig:wavenet}. 

\subsubsection{Dilated causal convolutions}
\label{dilations}
Standard convolutional networks cannot capture long temporal patterns with reasonably small models due to the increase in computational cost yielded by larger receptive fields. Dilated convolutions skip some input values so that the convolution kernel is applied over a larger area than its own. The network therefore operates on a larger scale, without the downside of increasing the number of parameters. The receptive field $r$ of a network made of stacked convolutions indeed reads: 
$$r = \sum_i d_i (s_i - 1),$$
where $d_i$ refers to the dilation rate ($d_i=1$ for normal convolutions) and $s_i$ the filter size of the $i^{th}$ layer. Additionally, \textit{causal} convolutions kernels ensure a causal ordering of input frames: the prediction emitted at time $t$ only depends on previous time stamps. It allows to reduce the latency at inference time.

\subsubsection{Gated activations and residual connections}
\begin{sloppypar}
As mentioned in \cite{van2016wavenet}, gated activations units -- a combination of tanh and sigmoid activations controlling the propagation of information to the next layer -- prove to efficiently model audio signals. Residual learning strategies such as skip connections are also introduced to speed up convergence and address the issue of vanishing gradients posed by the training of models of higher depth. Each layer yields two outputs: one is directly fed to the next layer as usual, but the second one skips it. All skip-connections outputs are then summed into the final output of the network. A large temporal dependency, can therefore be achieved by stacking multiple dilated convolution layers. By inserting residual connections between each layer, we are able to train a network of 24 layers on relatively small amount of data, which corresponds to a receptive field of 182 frames or 1.83s. The importance of gating and residual connections is analyzed in Section 3.3.2.
\end{sloppypar}

\subsection{Streaming inference}
\label{sec:streaming}
\begin{figure}[tb]
    \centering
    \includegraphics[width=7cm]{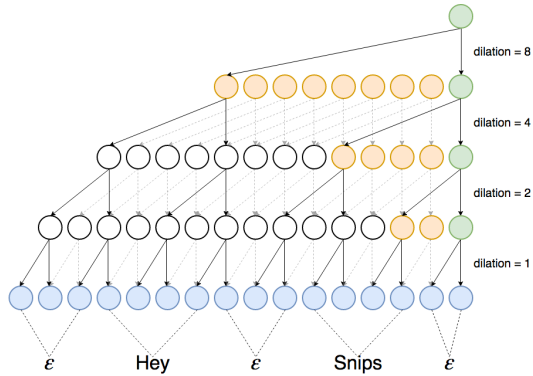}
    \caption{Dilated convolution layers with an exponential dilation rate of 1, 2, 4, 8 and filter size of 2. Blue nodes are input frame vectors, orange nodes are cached intermediate vectors used for streaming inference, green nodes are output vectors which are actually computed. $\epsilon$ refers to background.}
    \label{fig:dilations}
\end{figure}

In addition to reducing the model size, dilated convolutions allow the network to run in a streaming fashion during inference, drastically reducing the computational cost. When receiving a new input frame, the corresponding posteriors are recovered using previous computations, kept in memory for efficiency purposes as described in Figure~\ref{fig:dilations}. This cached implementation allows to reduce the amount of Floating Point Operations per Second (FLOPS) to a level suiting production requirements.

\subsection{End-of-keyword labeling}
\label{sec:eoh_labeling}
Our approach consists in associating a target 1 to frames within a given time interval $\Delta t$ before and after the end of the keyword. The optimal value for $\Delta t$ is tuned on the dev set. Additionally, a \textit{masking} scheme is applied, discarding background frames outside of the labeling window in positive samples. A traditional labeling approach, however, associates a target 1 to all frames aligned with the keyword. In this configuration, the model has a tendency to trigger as soon as the keyword starts, whether or not the sample contains only a fraction of the keyword. One advantage of our approach is that the network will trigger near the end of keyword, once it has seen enough context. Moreover, our labeling does not need any phoneme alignment, but only to detect the end of the keyword, which is easily obtained with a VAD system (only needed for labeling and not used for inference). Furthermore, thanks to masking, the precise frontiers of the labeling window are not learned, making the network more robust to labeling imprecisions. The relative importance of end-of-keyword labeling and masking are analyzed in Section \ref{sec:ablation}.

\section{Experiments}
\label{sec:exp}

\subsection{Open dataset}

The proposed approach is evaluated on a crowdsourced close-talk dataset. The chosen keyword is ``Hey Snips'' pronounced with no pause between the two words. The dataset contains a large variety of English accents and recording environments. Around 11K wake-word utterances and 86.5K ($\sim$96 hours) negative examples have been recorded, see Table~\ref{tab:data} for more details. Note that negative samples have been recorded in the same conditions than wake-word utterances, therefore arising from the same domain (speaker, hardware, environment, etc.). It thus prevents the model from discerning the two classes based on their domain-dependent acoustic features. 

\begin{table}[]
\centering
\begin{tabular}{@{}lcccc@{}}
\toprule
                    &                   & Train & Dev   & Test  \\ \midrule
                    & utterances           & 5876  & 2504  & 2588  \\
Hey Snips           & speakers       & 1179  & 516   & 520   \\
                    & max / speaker & 10  & 10  & 10  \\ \midrule
                    & utterances          & 45344 & 20321 & 20821 \\
Negative            & speakers       & 3330  & 1474  & 1469  \\
                    & max / speaker & 30 & 30 & 30 \\ \bottomrule
\end{tabular}
\caption{Dataset statistics.}
\label{tab:data}
\end{table}

Positive data has been cleaned by automatically removing samples of extreme duration, or samples with repeated occurrences of the wake word. Positive dev and test sets have been manually cleaned to discard any mispronunciations of the wake word (e.g. ``Hi Snips'' or ``Hey Snaips''), leaving the training set untouched. Noisy conditions are simulated by augmenting samples with music and noise background audio from Musan \cite{snyder2015musan}. The positive dev and test datasets are augmented at 5dB of Signal-to-noise Ratio (SNR).

The full dataset and its metadata are available for research purposes\footnote{https://research.snips.ai/datasets/keyword-spotting}. Although some keyword spotting datasets are freely available, such as the Speech Commands dataset \cite{warden2018speech} for voice commands classification, there is no equivalent in the specific wake-word detection field. By establishing an open reference for wake-word detection, we hope to contribute to promote transparency and reproducibility in a highly concurrent field where datasets are often kept private.

\subsection{Experimental setup}
The network consists in an initial causal convolution layer (filter size of 3) and 24 layers of gated dilated convolutions (filter size of 3). The 24 dilation rates are a repeating sequence of $\{1, 2, 4, 8, 1, 2, 4, 8...\}$. Residual connections are created between each layer and skip connections are accumulated at each layer and are eventually fed to a DNN followed by a softmax for classification as depicted in Figure~\ref{fig:wavenet}. We used projection layers of size 16 for residual connections and of size 32 for skip connections. The optimal duration of the end-of-keyword labeling interval as defined in Section \ref{sec:eoh_labeling} is $\Delta t = 160ms$ (15 frames before and 15 frames after the end of the keyword). The posteriors are smoothed over a sliding context window of $w_{smooth}=30$ frames, also tuned on the dev set.

The main baseline model is a LSTM trained with a max-pooling based loss initialized with a cross-entropy pre-trained network, as it is another example of end-to-end temporal model \cite{sun2016max}. 
The idea of the max-pooling loss is to teach the network to fire at its highest confidence time by back-propagating loss from the most informative keyword frame that has the maximum posterior for the corresponding keyword. 
More specifically, the network is a single layer of unidirectional LSTM with 128 memory blocks and a projection layer of dimension 64, following a similar configuration to \cite{sun2016max} but matching the same number of parameters than the proposed architecture (see Section \ref{sec:perf}). 10 frames in the past and 10 frames in the future are stacked to the input frame. Standard frame labeling is applied, but with the frame masking strategy described in Section \ref{sec:eoh_labeling}. The authors of \cite{sun2016max} mentioned back-propagating loss only from the last few frames, but said that the LSTM network performed poorly in this setting. The same smoothing strategy is applied on an window $w_{smooth}=8$ frames, after tuning on dev data. 
For comparison, we also add as a CNN variant the base architecture \texttt{trad-fpool3} from \cite{sainath2015convolutional}, a multi-class model with 4 output labels (``hey'', ``sni'', ``ps'', and background). Among those proposed in \cite{sainath2015convolutional}, this is the architecture with the lowest amount of FLOPS while having a similar number of parameters as the two other models studied here (see Section \ref{sec:perf}).

The Adam optimization method is used for the three models with a learning rate of $10^{-3}$ for the proposed architecture, $10^{-4}$ for the CNN, and $5 \cdot 10^{-5}$ for the LSTM baseline. Additionally, gradient norm clipping to 10 is applied. A scaled uniform distribution for initialization \cite{glorot2010understanding} (or ``Xavier'' initialization) yielded the best performance for the three models. We also note that the LSTM network is much more sensitive to the chosen initialization scheme.

\subsection{Results}

\subsubsection{System performance}
\label{sec:perf}

\begin{table}[]
\centering
\begin{tabular}{@{}l|cccc@{}}
\toprule
Model   & Params & FLOPS & FRR clean     & FRR noisy     \\ \midrule
WaveNet &  $222K$   &    $22 M$     & $0.12$ & $1.60$ \\
LSTM    &  $257K$   &    $26 M$     & $2.09$ & $11.21$   \\
CNN &  $244K$ & $172M$ & $2.51$ & $13.18$ \\ \bottomrule
\end{tabular}
\caption{Number of parameters, multiplications per second, and false rejection rate in percent on clean (FRR clean) and 5dB SNR noisy (FRR noisy) positive samples, at 0.5 false alarms per hour.}
\label{tab:frr}
\end{table}

\begin{figure}[htbp]
\centering
  \begin{subfigure}[t]{0.49\linewidth}
    \includegraphics[width=\textwidth]{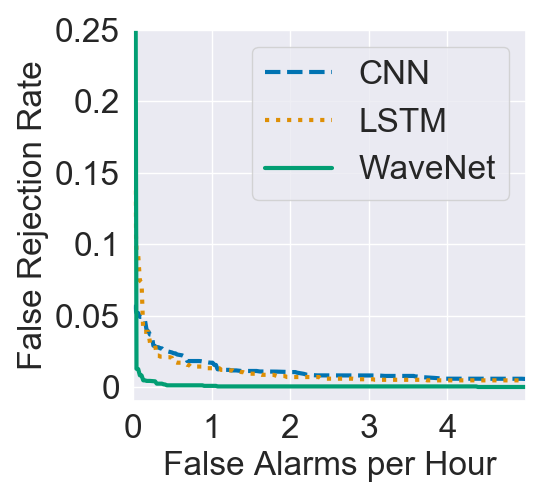}
    \vspace{-6mm}
    \caption{clean}
    \label{f1}
  \end{subfigure}
  \begin{subfigure}[t]{0.49\linewidth}
    \includegraphics[width=\textwidth]{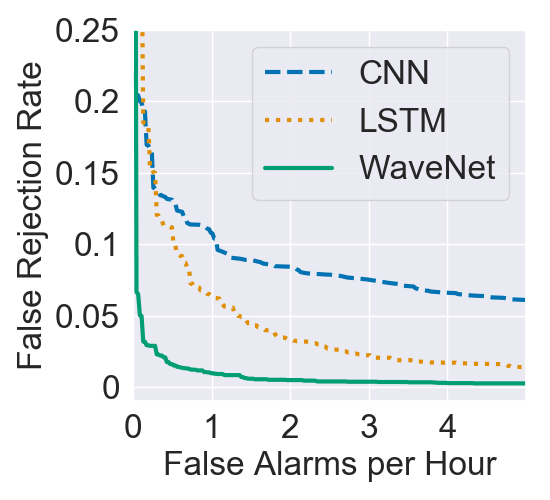}
    \vspace{-6mm}
    \caption{noisy (5dB SNR)}
    \label{f2}
  \end{subfigure}
\vspace{-3mm}
\caption{DET curves for the proposed architecture (green) compared to the LSTM (dotted yellow) and CNN (dashed blue) baselines in clean (a) and noisy (b) environments.}
\label{fig:roc}
\end{figure}

The performance of the three models is first measured by observing the False Rejection Rate (FRR) on clean and noisy (5dB SNR) positives samples at the operating threshold of 0.5 False Alarms per Hour (FAH) computed on the collected negative data. Hyper parameters are tuned on the dev set and results are reported on the test set. Table~\ref{tab:frr} displays these quantities as well as the number of parameters and multiplications per second performed during inference. The proposed architecture yields a lower FRR than the LSTM (resp. CNN) baseline with a 94\% (resp. 95\%) and 86\% (resp. 88\%) decrease in clean and noisy conditions. The number of parameters is similar for the three architectures, but the amount of FLOPS is higher by an order of magnitude for the CNN baseline while resulting in a poorer FRR in a noisy environment. Figure~\ref{fig:roc} provides the Detection Error Tradeoff (DET) curves and shows that the WaveNet model also outperforms the baselines on a whole range of triggering thresholds.


\subsubsection{Ablation analysis}
\label{sec:ablation}

To assess the relative importance of some characteristics of the proposed architecture, we study the difference in FRR observed once each of them is removed separately, all things being equal. Table~\ref{tab:ablation} shows that the end-of-keyword labeling is particularly helpful in improving the FRR at a fixed FAH, especially in noisy conditions. Masking background frames in positive samples also helps, but in a lower magnitude. Similarly to what is observed in \cite{chang2018temporal}, gating contributes to improving the FRR especially in noisy conditions. We finally observed that removing either residual or skip connections separately has little effect on the performance. However, we could not properly train the proposed model without any of these connections. It seems to confirm that implementing at least one bypassing strategy is key for constructing deeper network architectures.

\begin{table}[]
\centering
\begin{tabular}{@{}l|cc@{}}
\toprule
                 & FRR clean     & FRR noisy     \\ \midrule
Default labeling & $+0.36$ & $+1.33$ \\
No masking       & $+0.28$ & $+0.46$ \\
No gating  & $+0.24$ & $+2.57$ \\ \bottomrule
\end{tabular}
\caption{Variation in FRR (absolute) for the proposed architecture when removing different characteristics separately, all things being equal.}
\label{tab:ablation}
\end{table}

\section{Conclusion}
\label{sec:conclusion}

This paper introduces an end-to-end stateless modeling for keyword spotting, based on dilated convolutions coupled with residual connections and gating encouraged by the success of the WaveNet architecture in audio generation tasks \cite{van2016wavenet, chang2018temporal}. Additionally, a custom frame labeling is applied, associating a target 1 to frames located within a small time interval around the end of the keyword. The proposed architecture is compared against a LSTM baseline, similar to the one proposed in \cite{sun2016max}. Because of their binary targets, both the proposed model and the LSTM baseline do not require any phoneme alignment or post-processing besides posterior smoothing. We also added a multi-class CNN baseline \cite{sainath2015convolutional} for comparison. We have shown that the presented WaveNet model significantly reduces the false rejection rate at a fixed false alarm rate of 0.5 per hour, in both clean and noisy environments, on a crowdsourced dataset made publicly available for research purposes. The proposed model seems to be very efficient in the specific domain defined by this dataset and future work will focus on domain adaptation in terms of recording hardware, accents, or far-field settings, to be deployed easily in new environments.

\section{Acknowledgements}
We thank Oleksandr Olgashko for his contribution in developing the training framework. We are grateful to the crowd of contributors who recorded the dataset. We are indebted to the users of the Snips Voice Platform for valuable feedback.
\label{ssec:ack}

\bibliographystyle{plain}
\bibliography{strings,refs}

\end{document}